\documentclass[a4paper,12pt]{article}
\pdfoutput=1 
\usepackage{epsfig}
\usepackage{subfigure}
\usepackage{amsfonts}
\usepackage{times}
\usepackage{amsmath}
\usepackage{float}
\usepackage{tikz}  
\usetikzlibrary{automata}

\newcommand{\ben}{\begin{enumerate}}
\newcommand{\een}{\end{enumerate}}
\newcommand{\bi}{\begin{itemize}}
\newcommand{\ei}{\end{itemize}}
\newcommand{\bean}{\begin{eqnarray*}}          
\newcommand{\eean}{\end{eqnarray*}}
\newcommand{\bea}{\begin{eqnarray}}
\newcommand{\eea}{\end{eqnarray}}
\newcommand{\ba}{\begin{array}}
\newcommand{\ea}{\end{array}}
\newcommand{\bc}{\begin{center}}
\newcommand{\ec}{\end{center}}
\newcommand{\bt}{\begin{tabular}}
\newcommand{\et}{\end{tabular}}
\newcommand{\bfig}{\begin{figure}[htb]}
\newcommand{\efig}{\end{figure}}

\newcommand{\R}{\mathbb{R}}

\newcommand{\Gauss}[1]{\mathcal{N}(#1)}

\newcommand{\Ga}[1]{{\mathcal Ga}(#1)}

\newcommand{\eqnref}[1]{(\ref{#1})}
\newcommand{\figref}[1]{(\ref{#1})}

\title{\vspace*{-1cm} Constrained Mixture Models for Asset Returns Modelling}
\author{Iead Rezek$^1$ \vspace*{5mm}\\
Mathematical Imaging Neuroscience Group\\
Department of Neuroscience and Mental Health\\
Imperial College London, UK.
\thanks{Email: i.rezek@imperial.ac.uk}}

\date{\today}
\begin{document}

\maketitle
\vspace*{-1cm}

\abstract{The estimation of asset return distributions is crucial for determining optimal trading strategies. In this paper we describe the constrained mixture model, based on a mixture of Gamma and Gaussian distributions, to provide an accurate description of price trends as being clearly positive, negative or ranging while accounting for heavy tails and high kurtosis. The model is estimated in the Expectation Maximisation framework and model order estimation also respects the model's constraints.}

\tableofcontents
\section{INTRODUCTION}

The estimation of asset return distributions is crucial for determining optimal trading strategies.  One convenient estimation approach selects a distribution model and estimates its parameters. The advantage of this approach is  the ease with which probability distributions can be calibrated and applied in post-processing.  The disadvantage of assuming a particular parametric distribution is that inferences and decisions depend critically on the choice of distribution. For example, asset returns frequently feature large ``outlying'' values, making distributions with light tails inapplicable. 

Semi-parametric methods attempt to capture the advantages but not the disadvantages of 
a parametric specification of a returns distribution by using a more flexible functional form. Most prominent among the semi-parametric distributions are mixtures of 
distributions. They provide a flexible specification and, under certain conditions, can 
approximate distributions of any form.

\section{MIXTURE MODELS AND EXTENSIONS}

\subsection{Classical Mixture Models}

A standard mixture probability density of a random variable $X$,  whose value is denoted by $x$,  is defined as
\begin{equation}
p_{X}(x; \upsilon)=\sum_{k=1}^{K} \pi_{k} p_{X}(x;\theta_{k}).
\label{eqn:mixmodel}
\end{equation}
The mixture density has  $K$ components (or states) and is defined by the parameter set
$\upsilon=\{\theta,\pi \}$, where $\pi=\{\pi_{1},\cdots,\pi_{K}\}$ is the set of weights given to each component and  $\theta=\{\theta_{1},\cdots, \theta_{K}\}$ is the set of parameters describing each component distribution. 

By far, the most popular mixture model is the Gaussian mixture model (GMM). It is given as
\begin{equation}
p_{X}(x) =  \sum_{k=1}^{K} \pi_{k} \Gauss{x;\mu_{k},\sigma^{2}_{k}},
\end{equation}
where each component parameter vector $\theta_{k}$ now consists of the mean and variance parameters, $\mu_{k}$ and  $\sigma_{k}^{2}$, respectively (see Appendix~\ref{app:Standard Probability Distributions} for the definition of the probability distributions).

\begin{figure}[htbp] 
   \centering
   \includegraphics[height=.5\textheight,width=.9\textwidth]{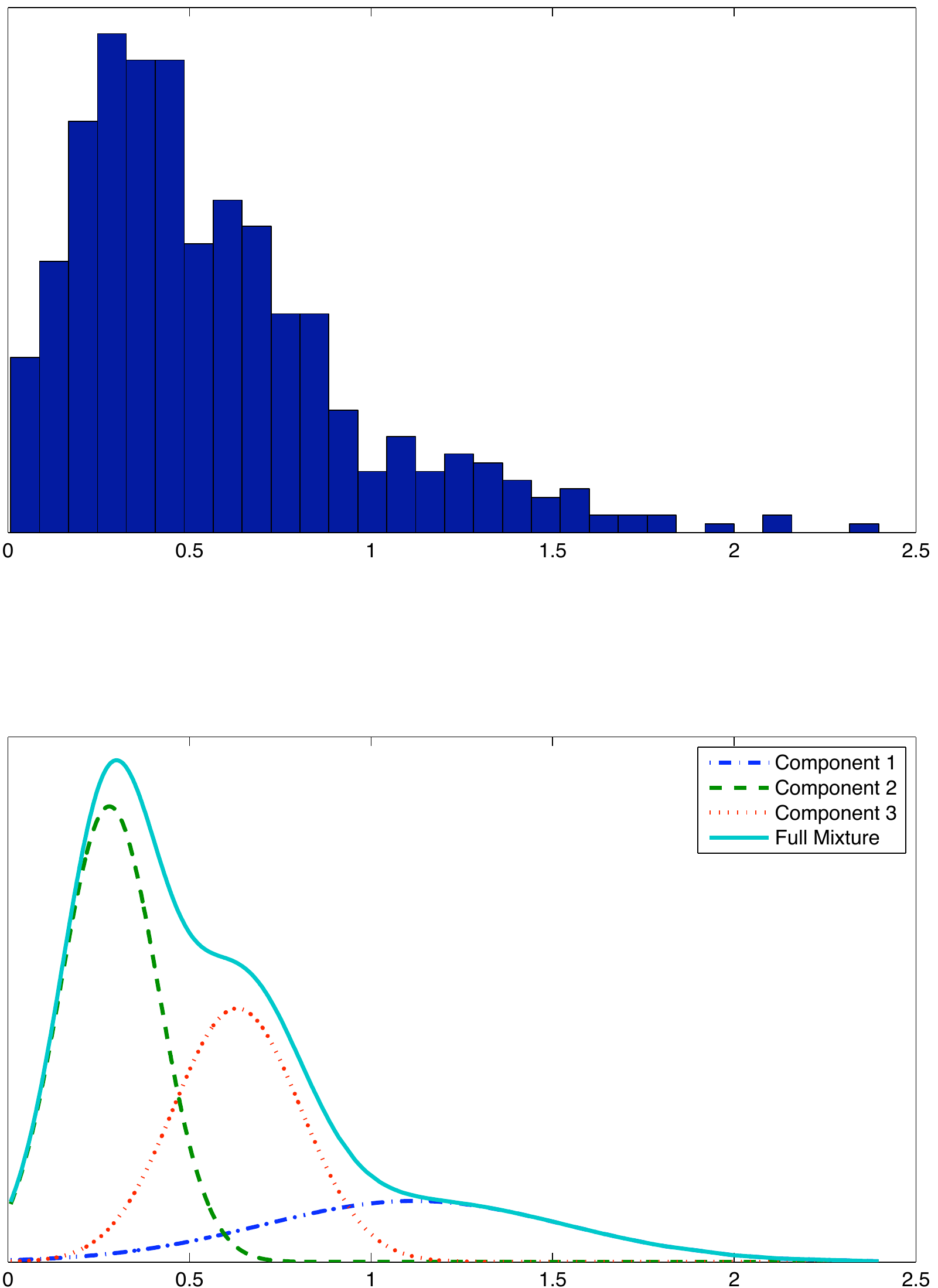} 
 \caption{{\it Histogram of a sample drawn randomly from a Gamma Distribution and the estimated Gaussian mixture model.}}
\label{fig:gammagaussmix}
\end{figure}

The Gaussian mixture distribution can be, and has been, estimated in the Maximum Likelihood or in a Bayesian framework (see~\cite{BishopCM:95} for both estimation methods). The Gaussian mixture distribution is often referred to as a universal approximator~\cite{BishopCM:95}, an indication of the fact that it can approximate distributions of any form. Figure~\figref{fig:gammagaussmix}, for example, shows a 3 component GMM approximating a sample with the histogram shown in the top plot. 

The number of components needed to model the data depends very much on the problem at hand. In some sense, it is the discrepancy between the data distribution and the mixture model that determines the number of components (aka model order). Data distributions with heavy tails require two or more light tailed components to compensate. In Figure~\figref{fig:gammagaussmix}, for example, the data was drawn from a single Gamma distribution yet  three Gaussian components were needed to capture most aspects of the Gamma distribution. 

More components require larger sample sizes to ensure adequate calibration. In the extreme case there may be insufficient data available to calibrate a given mixture model with a certain degree of accuracy. In short, while Gaussian mixture models are very flexible they may not be the most appropriate model.  If more is known about the data distribution, such as its behaviour in the tails, incorporation of this knowledge can only help improve the model.

\subsection{Gamma Mixture Models}

The Gamma mixture distribution is another commonly used model. They are used if the data values are only positive. Another reason for their use is because Gamma densities exhibit much heavier tails than Gaussian densities. Thus, events that deviate from the mean by several standard deviations are much more probable than under a Gaussian model assumption. As a consequence, large return values are not underestimated under the Gamma mixture assumption.

The Gamma mixture model (GaMM) is given as
\begin{equation}
p_{X}(x)=\sum_{k=1}^{K} \pi_{k} \Ga{X;\alpha_{k},\beta_{k}},
\end{equation}
where each component parameter vector $\theta_{k}$ now consists of the parameters shape and precision (inverse scale or rate), denoted respectively by $\alpha_{k}$ and  $\beta_{k}$(see Appendix~\ref{app:Standard Probability Distributions} for notation). 

The Gamma mixture distribution can be estimated via the Maximum Likelihood~\cite{LiuZ:2006} or the Bayesian framework~\cite{ChotikapanichD:2008}. Similar to its Gaussian counterpart, the Gamma mixture distribution can approximate any distribution on $\R^{+}$. 

Note that, for Bayesian inference, there is no natural prior for the shape parameter of the  Gamma distribution. Priors can be specified but require full MCMC (instead of Gibbs) sampling methods for estimation. With regard to maximum likelihood estimation note also that there is no closed form solution for the maximum likelihood estimator of the shape parameter - unless approximation assumptions are made~\cite{MinkaT:2002,LiuZ:2006} which then permit the use of gradient decent optimisation~\cite{MinkaT:2002}. Practice has shown, however, that even when making only small adjustments to the parameters the estimates frequently violate the positivity constraints, most notably that of the shape parameter.

Such limitations can be avoided, however, via the unique mapping that exists from the density's mean and variance to its shape and scale parameters
\begin{eqnarray}
\alpha &=& \frac{\mu}{\sigma^{2}} \nonumber\\
\beta &=&  \frac{\mu}{\sigma^{4}}
\label{eqn:gamparconv}
\end{eqnarray}
Thus, through the estimates resulting from the closed form solution of the mean and variance, the shape and scale parameter can be uniquely determined.

\section{Constrained Mixture Models}

Financial asset returns feature long positive and negative tails. In addition there is a large concentration of values around the origin. Modelling this constellation of distributions can be achieved by means of a Gaussian mixture model. However, as we pointed out earlier, heavy tail behaviour is more parsimoniously modelled with Gamma distributions. This fact leads to the obvious attempt to model large negative and positive values by Gamma distributions while a mixture of Gaussian densities takes on the task of modelling the sharply peaked distribution near the origin. This model is hereafter referred to as the constrained mixture model (CMM) or the Gauss-Gamma mixture distribution.

\subsection{Constraining by a Gauss-Gamma Mixture Distribution}

The main difference to standard mixture model is the association of subsets of components
$k=1,\cdots K$ to only positive and only negative valued observations. We will use the short hand notation $k\oplus$ for mixture component indices associated with positive observations. Likewise, $k\ominus$ refers  to the set of mixture component indices responsible for all negative valued observations. To specify the  remaining set of  component indices we use the symbol $k\odot$, i.e. $k\odot=\{1\cdots K\} \setminus \{ k\!\oplus \cup \, k\ominus\}$. For example, a $K=5$ component mixture model may be split into two components for positive valued observations $k\oplus=\{1,2\}$ and one  component for negative valued observations, $k\ominus=\{5\}$, whilst the remaining components are $k\odot=\{3,4\}$ and apply to all observations. 

The mixture component distributions are chosen according to which domain they are responsible for. We define three groups of mixture components as follows (see Appendix~\ref{app:Standard Probability Distributions} for notation):
\begin{description}
\item[Near Zero Domain:] Observations with values around zero are modelled by a set of Gaussian distributions which are {\em all} restricted to have zero mean. The probability of $x$ is thus
\begin{equation}
P_{X}(x ; \theta_{k})= \Gauss{x;\mu_{k}=0; \sigma^{2}_{k} } \;\;\;\;  \forall k\in k\odot
\end{equation} 
\item[Positive Domain:] Observations with positive values are modelled by a set of Gamma distributions. The probability of $x$ is thus
\begin{equation}
P_{X}(x | \theta_{k})= \Ga{x;\alpha_{k}; \beta_{k} } \;\;\;\;  \forall k\in k\oplus
\end{equation}
if the value $ x$ of $X$ is in $\R^{+}$ and zero, otherwise. 
\item[Negative Domain:] Observations with negative values are modelled by a set of Gamma distributions, and so the probability of $x$ is 
\begin{equation}
P_{X}(x | \theta_{k}) = \Ga{-x;\alpha_{k}; \beta_{k} } \;\;\;\;  \forall k\in k\ominus
\end{equation}
if the value $ x$ of $X$ is in $\R^{-}$ and zero, otherwise. 
\end{description}

Thus, the full constrained mixture model is given as 
\begin{equation}
p_{X}(x)=\left\{
\begin{array}{ccc}
\sum_{k\in k\oplus} \pi_{k} \Ga{x;\alpha_{k},\beta_{k}} &+ \sum_{ i \in k\odot} \pi_{i} \Gauss{X;0,\sigma^{2}_{i}} & \mbox{if } x\in \R_{0}^{+}\\
\sum_{k \in k\ominus}\pi_{k} \Ga{-x;\alpha_{k},\beta_{k}}&  +\sum_{ i \in k\odot} \pi_{i} \Gauss{x;0,\sigma^{2}_{i}} & \mbox{if }  x\in \R_{0}^{-}
\end{array}
\right.
\label{eqn:CMM}
\end{equation}

Note that negative values are modelled by a Gamma distribution with sign-reversed argument. In our notation, $k$ takes any of the index values of the states associated with constrained domain. A further consequence of our notation is that the parameter set $\theta$ consist of subsets $\theta_{k}$, each of which holds also the parameters of all other domains, e.g. $\theta_{3}=\{  \mu_{3}, \sigma^{2}_{3}, \alpha_{3}; \beta_{3}\}$. The reason for this is that we will be using the means and variances of the Gamma distribution to compute the distributions rate and scale parameter according to equations~\eqnref{eqn:gamparconv}.

\begin{figure}[htbp] 
   \centering
   \includegraphics[height=.35\textheight,width=\textwidth]{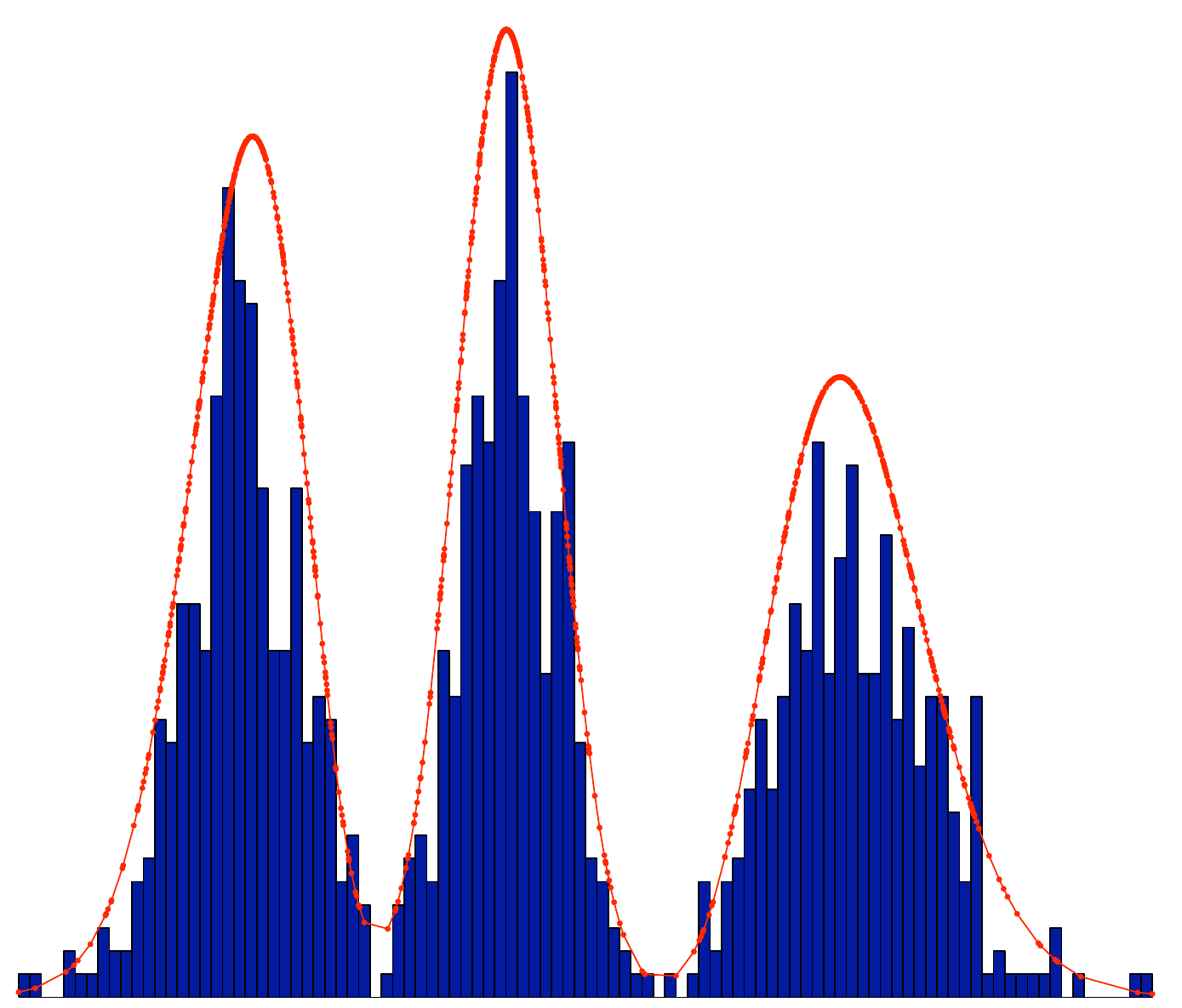} 
\caption{{\it Histogram of a sample drawn randomly from a constrained mixture model and the continuous density from which the sample was generated.}}
\label{fig:cmm}
\end{figure}

An example of a sample drawn from the constrained mixture and it's continuous density function are shown in Figure~\figref{fig:cmm}.
	
\subsection{Alternative Approaches to Constraining Distributions}

There are other ways to constrain the model. One way would be through the use of rectified Gaussian distributions~\cite{WinnJ:2004,HarvaM:2007}. However, the models in
~\cite{HarvaM:2007} use a cut-off function
\begin{equation}
\mbox{cut}(x) = \max(x,0) 
\end{equation}
which places too much weight on zero. Also,the CMM is considerably simpler while perfectly satisfying the required constraints.

\section{Mixture Model Estimation}

To motivate the estimation procedure we need to expand the mixture model. In particular, we introduce, for each datum, a latent indicator variable. This variable indicates which of the mixture component is responsible  for the datum in question. The (marginal) distribution that any indicator variable selects the $k$-th component is given by the weight $\pi_{k}$ that is associated with the $k$-th mixture component.

\subsection{Latent Indicator Variable Representation of Mixture Models}


Let us first define the following one-dimensional observation set 
$X=\{X_{1},\cdots,X_t,\cdots X_{T}\}$, of length $T$ and indexed with $t$. The set is assumed to be generated by a  $K$-component mixture model.

To indicate the mixture component from which a sample was drawn, we introduce a latent random variable, $S_{t}$. The value of $S_{t}$, which we denote by $s_{t}$, is a vector of length $K$. The components of the vector, $s_{t_{k}}$ are either $0$ or $1$.  We set the vector's $k$-th component, $s_{t_{k}}=1$  to indicate that the $k$-th mixture component is selected, while all other states are set to $0$. As a consequence, 
\begin{equation}
1=\sum_{k=1}^{K} s_{t_{k}}.
\end{equation}

We can now specify the joint probability distribution of $X$ and  $S$ in terms of a  marginal distribution $P_{S_{t}}(s_{t};\pi) $ and a conditional distribution $P_{X_{t}|S_{t}}(x_{t} |s_{t} ;\theta)$ as
\begin{eqnarray}
P_{X,S}(x,s;\upsilon) =   \prod_{t}^{T}P_{X_{t} |S_{t}}(x_{t} |s_{t} ;\theta) P_{S_{t}}(s_{t};\pi) ,
\label{eqn:completeprob}
\end{eqnarray}
and where the parameter vector $\upsilon=\{\theta,\pi\}$.

The marginal distribution $P_{S_{t}}(s_{t};\pi)$  are drawn from a multinomial distribution that is parameterised by the mixing weights $\pi=\{\pi_{1}\cdots\pi_{K}\}$. Thus,
 \begin{equation}
P_{S_{t}}(s_{t};\pi)= \prod_{k=1}^{K} \pi_{k}^{s_{t_{k}}}
\end{equation}
or, more simply,
\begin{equation}
P(s_{t_{k}}=1) =\pi_{k}.
\end{equation}
Naturally the weights must satisfy that $\pi_{k} \in [0, 1]$ and that
\begin{equation}
1=\sum_{k=1}^{K} \pi_{k}.
\end{equation}

As for the conditional distribution, $P_{X_{t} |S_{t}}(x_{t} |s_{t} ;\theta)$, its form depends on the value of the latent variable $S_{t}$. For the constrained mixture model we have in particular
\begin{equation}
P_{X_{t}}(x_{t} | s_{t_{k}}=1; \theta_{k})=\left\{
\begin{array}{cccc}
 \Gauss{x_{t};0; \sigma^{2}_{k} }  \; & \forall x_{t}  & \mbox{and}  & k \in {k\odot}\\
\Ga{x_{t} ;\alpha_{k}; \beta_{k} }   \;\;\; & x_{t}  \in \R^{+} & \mbox{and} & k \in {k\oplus}\\
\Ga{-x_{t} ;\alpha_{k}; \beta_{k} }  & x_{t} \in \R^{-} & \mbox{and} & k \in {k\ominus} \\
0 & \mbox{otherwise}
\end{array}\right. 
\end{equation}

The full model is thus defined as
\begin{equation}
P_{X,S}(x,s;\upsilon)  = \prod_{t}^{T} \prod_{k=1}^{K} \pi_{k}^{s_{t_{k}}} \left\{
\begin{array}{cccc}
\Gauss{x_{t};0; \sigma^{2}_{k} }  \;    & k \in {k\odot} & \mbox{and} & \forall x_{t}\\
\Ga{x_{t} ;\alpha_{k}; \beta_{k} }   \;\;\;  &  k \in {k\oplus} & \mbox{and} & x_{t}  \in \R^{+}\\
\Ga{-x_{t} ;\alpha_{k}; \beta_{k} } & k \in {k\ominus}  & \mbox{and} & x_{t} \in \R^{-} \\
0 & \mbox{otherwise}
\end{array}\right. 
\label{eqn:fullmodel}
\end{equation}

To summarise, in  the latent  variable representation of mixture model, the components for each sample are selected with probability $\pi_{k}$ $\forall k$, reflecting the mixture weight $\pi_{k}$. The components that are selected for a particular datum $x_{t}$ depend on the sign of the sample $x_{t}$. For positive $x$, $x_{t}$ is modelled by a mixture of Gaussian and ``positive'' Gamma distributions. For negative $x_{t}$, $x_{t}$ is modelled by the same mixture of Gaussian and a set of ``negative'' Gamma distributions.

\subsection{Maximum Likelihood Estimation}

Estimation of the mixture model can be accomplished by maximising directly the model  given by equation~\eqnref{eqn:CMM}. This, however, requires the use of optimisation methods such as the Newton-Raphson algorithm. Using the complete data mixture model description instead leads to an optimisation algorithm known as the Expectation-Maximisation algorithm. The algorithm produces set of coupled yet analytic update equations that can be iterated until convergence has been achieved. What is more, convergence is easily monitored since the convergence criterion is simply one of the quantities that the algorithm computes anyhow.

The maximum likelihood method of estimating mixture models used here is known as the Expectation Maximisation (EM) algorithm. The goal of the EM is to maximise the likelihood of the data given the model, i.e. maximise 
\begin{equation}
{\cal L}(\upsilon) = \log\left\{ \sum_{s} P_{X,S}(x,s;\upsilon) \right\} =
 \sum_{t=1}^{T}   \sum_{k=1}^{K} s_{t_{k}}\log\left\{ \pi_{k}  P_{X_{t}}(x_{t}; \theta_{k}) \right\}
 \label{eqn:fullloglike} 
\end{equation}

If the states of  $S=\{S_{0},S_{1},\cdots,S_t,\cdots S_{T}\}$ had been known then the estimation of the model parameters $\pi,\theta$ is trivial. Conditioned on the 
state variables and the observations, the equation~\eqnref{eqn:fullloglike} could be maximised with respect to the  model parameters. However, which value that the state variables take is unknown. This suggests an alternative two-stage iterated optimisation algorithm: If we know the   \emph{expected} of $S$, one could use this expectation in the first step to perform a weighted maximum likelihood estimation of~\eqnref{eqn:fullloglike} with respect to the model parameters. These estimates will be incorrect since the expectation $S$ is inaccurate. So, in the second step, one could update the expected value of all $S$ subject to the pretending the model parameters $\pi$ and $\theta$ are known and held fixed at their values from the past iteration. This is precisely the strategy of the Expectation Maximisation (EM) algorithm~\cite{BishopCM:95}.
  
The EM algorithm for the CMM iteratively optimises  ${\cal L}(\upsilon)$ in two stages~\cite{BishopCM:95}:
\begin{description}
\item[E-step] In this step, the parameters $\upsilon$ are held fixed at the old values,  
$\upsilon^{old}$,  obtained from the previous iteration (or at their initial settings during the algorithm's initialisation). Conditioned on the observations,  the E-step then computes the probability of the state variables $S_{t}$, $\forall t$ given the current model parameters and observation data, i.e.
\begin{equation}
P_{S_{t}|X_{t}}(s_{t}|x_{t},\upsilon^{old}) \propto P_{X_{t}|S_{t}}(x_{t}|s_{t};\theta) P_{S_{t}}(s_{t};\pi^{old})
\end{equation}
In particular, we compute (and drop the superscript for clarity's sake)
\begin{equation}
P_{S_{t}|X_{t}}(s_{t_{k}}=1|x_{t},\upsilon^{old})=\frac{P_{X_{t}|S_{t}}(x_{t}|s_{t_{k}}=1;\theta_{k}) \pi_{k}}{\sum_{s_{t_{\ell}}}P_{X_{t}|S_{t}}(x_{t}|s_{t_{\ell}}=1;\theta_{\ell}) \pi_{\ell}}
\label{eqn:posterior}
\end{equation}
The likelihood terms $P_{X_{t}|S_{t}}(x_{t}|s_{t_{k}}=1;\theta_{k})$ are evaluate using the  observation densities defined for each of the states. Thus,
\begin{equation}
P_{X_{t}|S_{t}}(x_{t}|s_{t_{k}}=1;\theta_{k})=\left\{
\begin{array}{cccc}
\Gauss{x_{t};0; \sigma^{2}_{k} }  \;    & k \in {k\odot} & \mbox{and} & \forall x_{t}\\
\Ga{x_{t} ;\alpha_{k}; \beta_{k} }   \;\;\;  &  k \in {k\oplus} & \mbox{and} & x_{t}  \in \R^{+}\\
\Ga{-x_{t} ;\alpha_{k}; \beta_{k} } & k \in {k\ominus}  & \mbox{and} & x_{t} \in \R^{-} \\
0 & \mbox{otherwise}
\end{array}\right. 
\end{equation}
To simplify the notation we use $\gamma_{t}$ to symbolise the vector values computed in~\eqnref{eqn:posterior}, which are the probabilities for each component $k$ being selected for observation $x_{t}$. The components of $\gamma_{t}$ are denoted by $\gamma_{t_{k}}$, i.e. 
\begin{equation}
\gamma_{t_{k}} = P_{S_{t} | X_{t}}(s_{t_{k}}=1 | x_{t};\upsilon^{old}). 
\end{equation}
Note that, as a consequence of equation~\eqnref{eqn:posterior}, $1=\sum_{k=1}^{K}\gamma_{t_{k}}$. 
\item[M-step] In this step, the latent state probabilities are considered given and maximisation is performed with respect to  the parameters $\theta$:
\begin{equation}
\upsilon^{new}=\arg\max_{\upsilon} {\cal L}(\upsilon)
\end{equation}
This results in the update equations for the parameters for the probability distributions are as follows
\begin{eqnarray}
\mu_{k}&=&\frac{1}{T} \sum_{t=1}^{T} \gamma_{t_{k}} x_{t}\\
\sigma^{2}_{k}&=&\frac{1}{T} \sum_{t=1}^{T} \gamma_{t_{k}}) \left(x_{t} - \mu_{k}\right)^{2}
\end{eqnarray}
These two parameters are computed for all states. 

For those states that are governed by a Gamma distribution, the shape and scale parameters are computed using the relations
\begin{eqnarray}
\alpha_{k} &=& \frac{\mu_{k}}{\sigma^{2}_{k}}\\
\beta_{k} &=&  \frac{\mu_{k}}{\sigma^{4}_{k}}
\end{eqnarray}
This approach circumvents the need for approximations or an iterative gradient decent approach to optimising the shape parameter $\alpha_{k}$ .
\end{description}

\section{Results}

Before applying the model to some data it is worth studying the  model and the training  algorithm's behaviour on a simulated data set.
    
\begin{figure}[H] 
   \centering
   \includegraphics[height=.4\textheight,width=.9\textwidth]{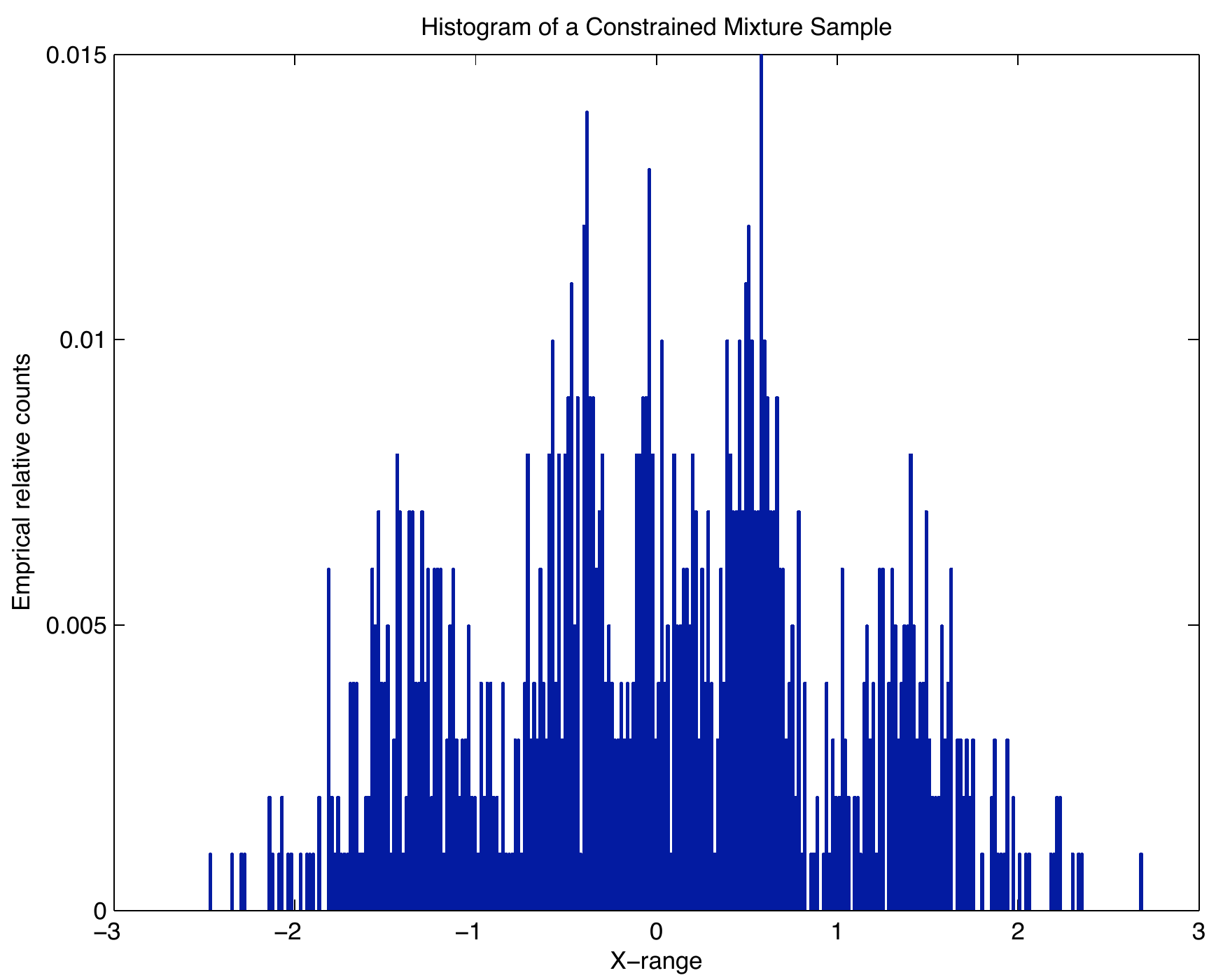} 
   \caption{Histogram of a sample drawn from a constrained mixture model with 2 Gamma 
   distributions for the positive x-values, 2 Gamma  distributions for the negative x-values
   and one Gaussian distribution centred at the origin.}
   \label{fig:cmm_hist}
\end{figure}

\subsection{Simulated Results}

We generated data from a pre-specified constrained mixture model. In the model, there were  2 Gamma distributions assigned to the positive domain. These had, respectively, the shape parameters $\alpha_{p1}=20$ and $\alpha_{p2}=10$ and scale parameters $\beta_{p1}=3$ and $\beta_{p2}=4$.  Assigned to the negative domain were also by 2 Gamma distributions. Respectively, their  shape parameters are $\alpha_{n1}=20$ and $\alpha_{n2}=10$ and scale parameters $\beta_{n1}=3$ and $\beta_{n2}=4$. Finally, a single Gaussian distribution was also defined to be centred at the origin with a variance of $\sigma_{0}^{2}=1$.  

\begin{figure}[htbp] 
    \centering
    \includegraphics[height=.25\textheight,width=.9\textwidth]{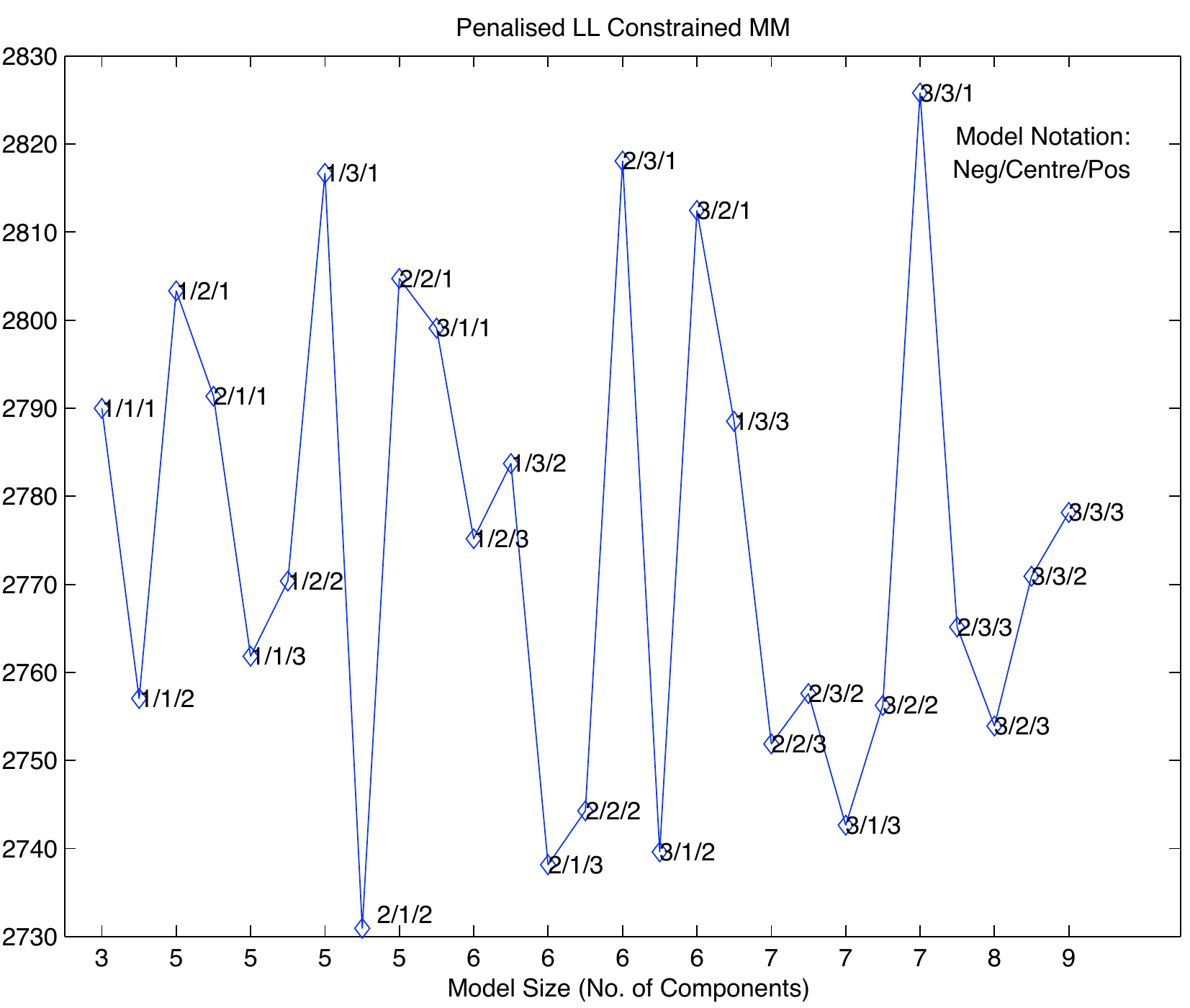} 
    \caption{Penalised likelihood values for 27 configurations of the contrained mixture model. Each model is denoted by  $k\!\ominus\!/k\!\odot\!/k\oplus$, a system of numbers indicating that there are $k\ominus$ Gamma distributions defined for the negative domain, $k\oplus$ Gamma distributions defined for the positive domain and $k\odot$ Gaussian distributions centred at the origin.}
    \label{fig:cmm_mo}
 \end{figure}

\begin{figure}[htbp] 
   \centering
   \includegraphics[height=.25\textheight,width=.8\textwidth]{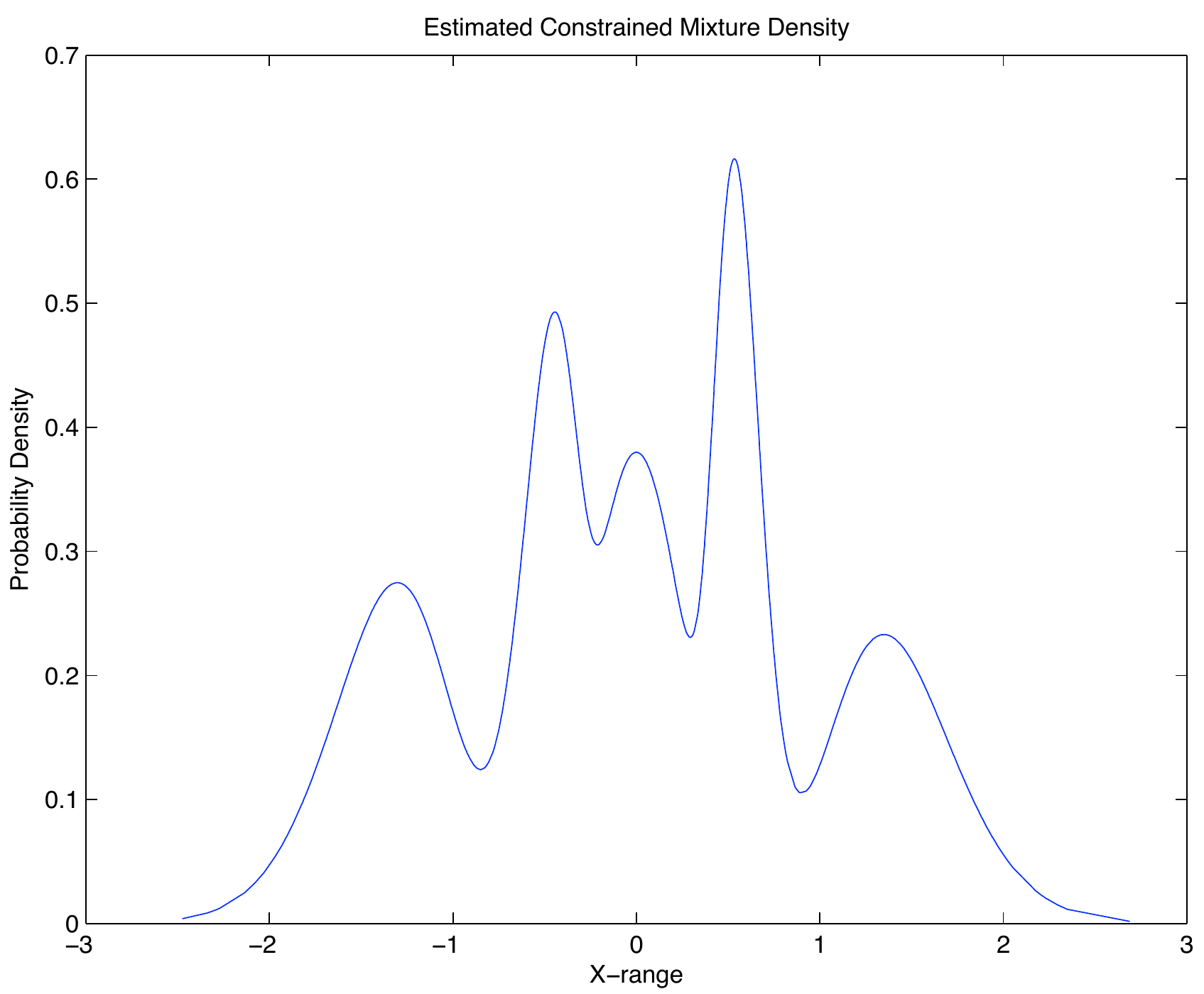} 
   \caption{The estimated density for a sample drawn from the $2/1/2$-constrained mixture model described in the text.}
   \label{fig:cmm_est}
\end{figure}

A total of $1000$ samples were drawn from the constrained distribution. The empirical relative counts, i.e. the histogram, is shown in Figure~\figref{fig:cmm_hist}. Model calibration was subsequently repeated for a range for model orders. In particular, the number of kernels for the negative values ranged from $1-3$, similarly for the positive values and the centred Gaussians. Thus a total of $27$ model configurations were evaluated. The penalised likelihood (BIC~\cite{BishopCM:95}) for each model is shown in Figure~\figref{fig:cmm_mo}.  The minimum penalised likelihood, i.e. the most parsimonious, configuration was found for precisely the configuration from which the data was sampled ($2$ negative Gamma p.d.f.s, $2$ positive Gamma p.d.f.s, and $1$ Gaussian p.d.f.). The resultant estimated constrained model is shown in Figure~\figref{fig:cmm_est}.

A number of things are noteworthy. The total number of $27$ model configurations implies a large number of computations. This is due to constrained nature of the model. These computations are not necessary in that it is similarly possible to estimate the total number of mixture components using a standard Gaussian mixture model, which in this example would imply maximally $9$ components. The allocation of kernels to domains in the constrained mixture can then be determined through visual inspection of the fitted Gaussian mixture. This approach is approximately statistically correct. The implied assumption is that each of the Gamma distributions is sufficiently accurately fitted by a Gaussian distribution. 

Penalising the log-likelihood using BIC, or any other off-the-shelf penalty term, is theoretically incorrect. This is due to the fact that standard penalty criteria assume that all model parameters are used to explain the same number of samples - as expressed by $p\log T$ in the BIC case, $p$ being the number of model parameters and $T$ the sample size. This condition does not apply in the constrained mixture model case. Gamma distributions are only used to fit samples that fall within their domain of responsibility. While it is possible to  modify the  penalty criteria to match the constrained model, the standard penalty factors suffice in practice. The standard penalty factors are at worst overly conservative, i.e. the recommend model order is smaller than one obtained by a constrained-model matching criterion. 

\subsection{Asset Returns}

\begin{figure}[H] 
   \centering
   \includegraphics[height=.4\textheight,width=.9\textwidth]{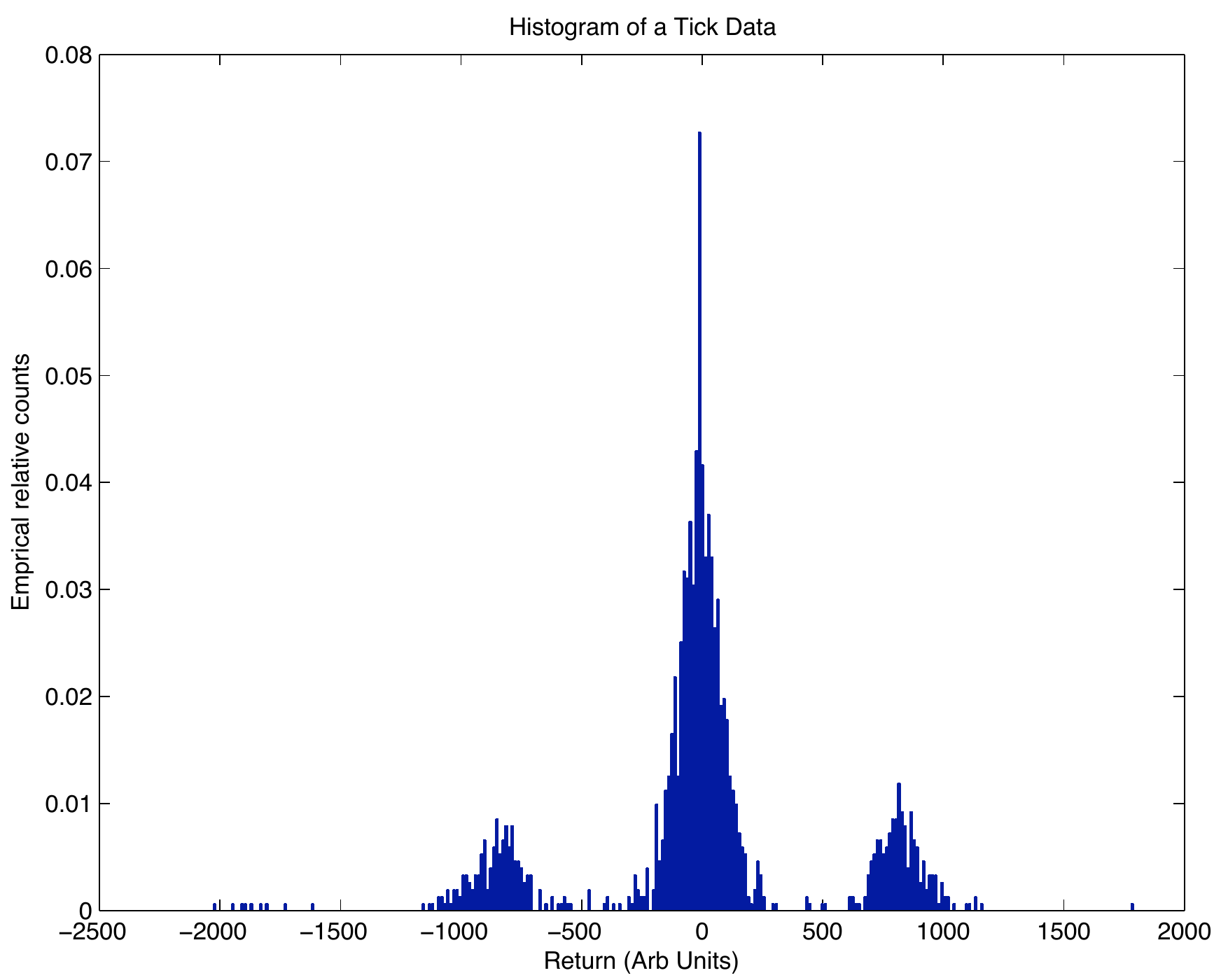} 
   \caption{Histogram of a sample of $1513$ ticks obtained from difference of the average daily price of the US Treasury $10$-year bonds.}
   \label{fig:cmmtick_hist}
\end{figure}

We now describe the application of the model (and the model order selection via penalised likelihood) to actual financial data. The data is the US Treasury $10$-year bond price, collected over the period of $5$ years on a daily basis - exactly $1513$ trading days. The asset's returns were calculated as the difference of the day's average price from that of the previous day. The sample's histogram is shown in Figure~\figref{fig:cmmtick_hist}.
 
\begin{figure}[H] 
   \centering
   \includegraphics[height=.4\textheight,width=.9\textwidth]{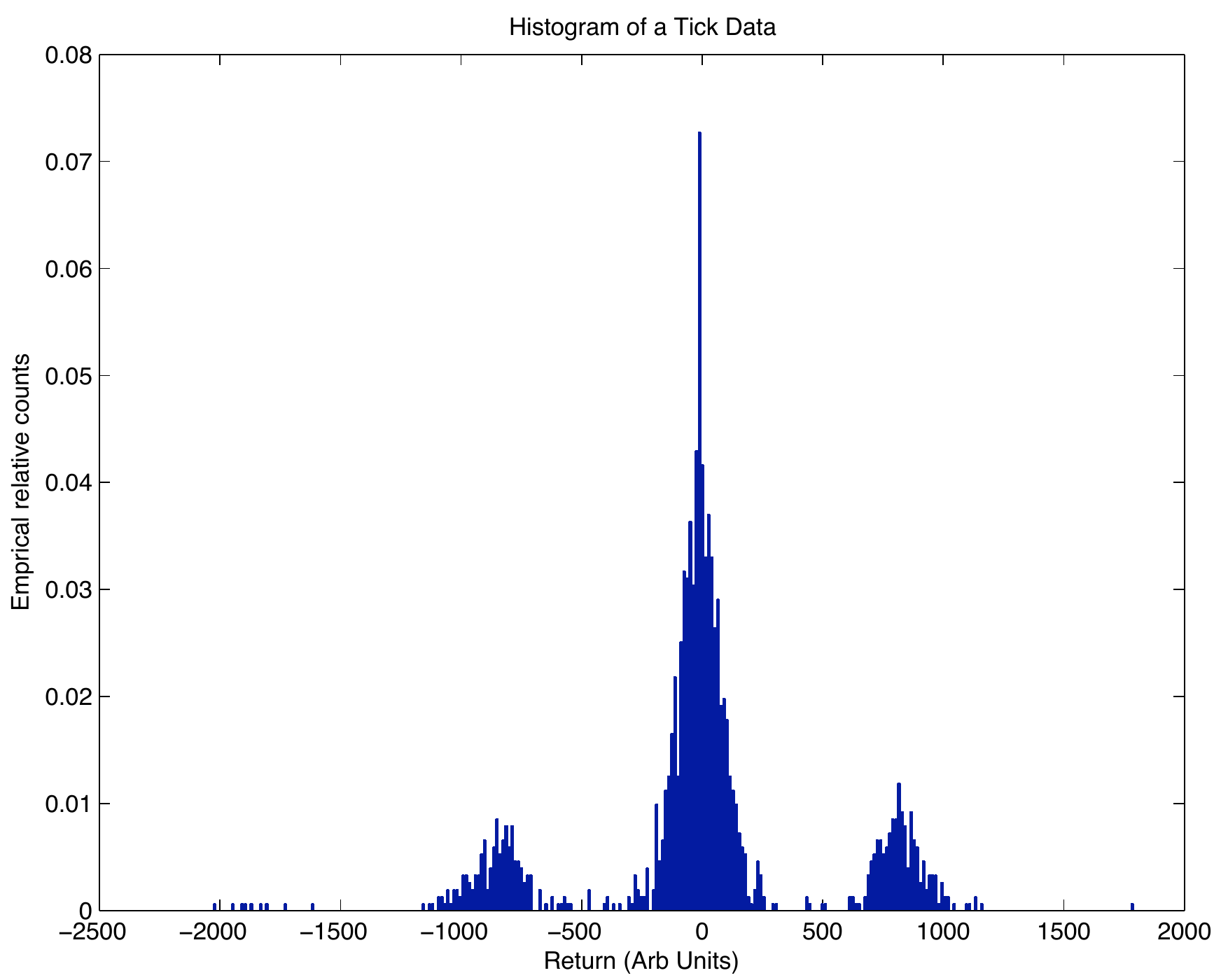} 
   \caption{The estimated density for the difference in average daily price of the US Treasury $10$-year bonds, using a $1/1/1$-constrained mixture model.}
   \label{fig:cmmtick_est}
\end{figure}

The optimal model order that was determined using maximum likelihood estimation and penalising using the BIC penalty criterion. The configuration thus calculated was $1/1/1$, i.e. 
$1$ Gamma distributions defined for the negative domain, $1$ Gaussian distributions centred at the origin and $1$ Gamma distributions defined for the positive domain. The resulting mixture model fit is shown in Figure~\figref{fig:cmmtick_est} and suggest a good fit.
 
\section{Discussion}

The constrained mixture model provides a simple statistical decomposition into negative, positive and near zero domains. The motivation for this model is the accurate description of  price trends as being clearly positive, negative or ranging while accounting for heavy tails and high kurtosis.  

The EM algorithm for the constrained mixture model is only marginally different from that of standard mixture models. Model estimation can be performed using standard likelihood penalisation methods. Even though theoretically over-penalising, the study on simulated data has shown that their use does produce an acceptable model complexity estimates.  

Issues that remain to be solved are largely identifiability issues. As an example, a Gaussian distribution at the centre, flanked by two identical Gamma distributions provide as good a model as one where the two Gamma distributions are replace by one or two Gaussian distributions. While it is of theoretical concern and may imply increase sensitivity to the initialisations of the model parameters, in practice, such precise symmetry may never arise.


\clearpage

\bibliographystyle{unsrt}
\bibliography{biblist}

\appendix

\section{Standard Probability Distributions}
\label{app:Standard Probability Distributions}

\subsection{The Normal or Gaussian Probability Density}
The Normal probability density, denoted by $\Gauss{x;\mu,\sigma^{2}}$, is given as
\begin{equation}
P_{X}(x) = \frac{1}{\sqrt{2 \pi \sigma^{2}}} e^{ -\frac{1}{2 \sigma^{2}} \left(x-\mu\right)^{2 } }
\end{equation}
where  $\mu$ is the mean and the variance is $\sigma^{2}$. 

\subsection{The Gamma Distribution}

The Gamma probability density, denoted by $\Ga{x;\alpha,\beta}$, is given as
\begin{equation}
P_{X}(x)  = \frac{\beta^{\alpha}}{\Gamma(\alpha)} x^{\alpha-1} e^{-\beta x}
\end{equation}
where  $\alpha$ is the shape parameters and $\beta$ is the inverse scale (or rate).

\end{document}